\documentclass{article} 
\usepackage[table,xcdraw]{xcolor}
\usepackage{iclr2024_conference,times}
\usepackage{multirow}
\usepackage{bbding}

\usepackage{amsmath,amsfonts,bm}

\def\eqref#1{equation~\ref{#1}}

\def\1{\bm{1}}

\DeclareMathAlphabet{\mathsfit}{\encodingdefault}{\sfdefault}{m}{sl}
\SetMathAlphabet{\mathsfit}{bold}{\encodingdefault}{\sfdefault}{bx}{n}

\usepackage{hyperref}       
\hypersetup{
    colorlinks = true,
    linkcolor = magenta,
    anchorcolor = blue,
    citecolor = cyan,
    filecolor = magenta,
    urlcolor = magenta,
    pdftitle = {Language Model Composition},
}

\usepackage{url}
\usepackage{booktabs} 
\usepackage{algorithm}
\usepackage{algpseudocode}
\usepackage{adjustbox}
\usepackage{multicol}
\usepackage{wrapfig}
\usepackage{url,xspace}
\usepackage{graphicx}
\usepackage[labelformat=empty, skip=0pt]{subcaption}
\usepackage{cleveref}
\usepackage{listings}
\usepackage{tabularx}
\usepackage{array}
\usepackage{listliketab}
\usepackage{tikz}
\usepackage{arydshln}

\usepackage{color}

\usepackage{listings}

\crefformat{section}{\S#2#1#3} 
\crefformat{subsection}{\S#2#1#3}
\crefformat{subsubsection}{\S#2#1#3}

\DeclareFixedFont{\ttb}{T1}{txtt}{bx}{n}{8} 
\DeclareFixedFont{\ttm}{T1}{txtt}{m}{n}{8}  

\newcommand{\fullName}{Composition to Augment Language Models}
\newcommand{\name}{CALM}

\newcommand{\model}{$\mathbf{m}$}
\newcommand{\layerSet}{$\mathbb{L}$}

\newcommand{\layerIdxT}[2]{\text{#1}#2}

\newcommand{\repVectorSet}{$\mathbb{H}$}

\newcommand{\repVectorT}{\mathbf{H}}
\newcommand{\projLayer}{\mathnormal{f_{\text{proj}}}}
\newcommand{\crossAttn}{\mathnormal{f_{\text{cross}}}}
\newcommand{\projLayerT}[2]{\projLayer(\repVectorT_{\text{#1}#2})}

\newcommand{\crossAttnT}[4]{\crossAttn(\projLayerT{#1}{#2}, \repVectorT_{\text{#3}#4})}

\newcommand{\composeData}[1]{$\mathbf{D}_\mathbf{C}^{\text{#1}}$}

\newcommand{\modelT}[1]{\model$_{\text{#1}}$}
\newcommand{\aug}{A}
\newcommand{\anchor}{B}
\newcommand{\augM}{\modelT{\aug}}
\newcommand{\anchorM}{\modelT{\anchor}}

\newcommand{\composed}[2]{\model$_{\text{\aug}\oplus\text{\anchor}}$}
\newcommand{\composedNot}[2]{\modelT{C}}
\newcommand{\finetuned}[2]{\model$_{\text{#1}}^{\text{#2}}$}
\newcommand{\knowledgeData}[1]{$\mathbf{D}_{\text{#1}}$}

\newcommand{\palmxxs}{{PaLM2-XXS}}
\newcommand{\palmxs}{{PaLM2-XS}}
\newcommand{\palms}{{PaLM2-S}}

\definecolor{deepblue}{rgb}{0,0,0.5}
\definecolor{deepred}{rgb}{0.6,0,0}
\definecolor{deepgreen}{rgb}{0,0.5,0}

\definecolor{baseColorT}{HTML}{FFFFFF}
\definecolor{baselineColorT}{HTML}{EFEFEF}
\definecolor{smallColorT}{HTML}{fff1e6}
\definecolor{largeColorT}{HTML}{edf2fb}
\definecolor{composeColorT}{HTML}{f8f7ff}

\definecolor{baseColor}{HTML}{EFEFEF}
\definecolor{smallColor}{HTML}{CA8634}
\definecolor{largeColor}{HTML}{3473CA}
\definecolor{composeColor}{HTML}{907BB8}

\definecolor{dullRed}{HTML}{FCEFEB}
\definecolor{dullGreen}{HTML}{F5FCEB}
\definecolor{dullGray}{HTML}{F3F3F3}

\newcommand{\customdashline}{\noalign{\vskip 0.2ex} \hdashline \noalign{\vskip 0.5ex}} 

\newcommand\pythonstyle{\lstset{
language=Python,
basicstyle=\ttm,
morekeywords={self},              
keywordstyle=\ttb\color{deepblue},
emph={MyClass,__init__},          
emphstyle=\ttb\color{deepred},    
stringstyle=\color{deepgreen},
showstringspaces=false
}}

\lstnewenvironment{python}[1][]
{
\pythonstyle
\lstset{#1}
}
{}

\newcommand\pythoninline[1]{{\pythonstyle\lstinline!#1!}}

\newcommand{\rac}[1]{}
\newcommand{\bid}[1]{}
\newcommand{\sid}[1]{}
\newcommand{\nit}[1]{}
\newcommand{\pa}[1]{}
\newcommand{\pra}[1]{}

\newcommand{\blfootnote}[1]{{\renewcommand{\thefootnote}{\roman{footnote}}\footnotetext[0]{#1}}}

\newcommand*\emptysubfigure[1]{\begin{subfigure}[]{0pt}\caption{}\label{#1}\end{subfigure}}

\title{LLM Augmented LLMs:\\
Expanding Capabilities through Composition
}

\author{Rachit Bansal$^1$ Bidisha Samanta$^1$ Siddharth Dalmia$^2$ Nitish Gupta$^1$ Shikhar Vashishth$^1$ \\ \textbf{Sriram Ganapathy}$^1$ \textbf{Abhishek Bapna}$^1$ \textbf{Prateek Jain}$^1$ \textbf{Partha Talukdar}$^1$ \\
$^1$Google Research\quad$^2$Google DeepMind
}

\iclrfinalcopy 
\begin{document}

\maketitle

\blfootnote{
Correspondence to Rachit and Bidisha: [\href{mailto:brachit@google.com}{brachit}, \href{mailto:bidishasamanta@google.com}{bidishasamanta}]@google.com
}

\begin{abstract}

Foundational models with billions of parameters 
which have been trained on large corpora of data have demonstrated non-trivial skills in a variety of domains.
However, due to their monolithic structure, it is challenging and expensive to augment them or impart new skills.
On the other hand, due to their adaptation abilities, several new instances of these models are being trained towards new domains and tasks.
In this work, we study the problem of efficient and practical composition of existing foundation models with more specific models to enable newer capabilities.
To this end, we propose \name{}---\fullName{}---which introduces
cross-attention between models to compose their representations and enable new capabilities.
Salient features of \name\ are: (i) Scales up LLMs on new tasks by `re-using' existing LLMs along with a few  additional parameters and data,
(ii) Existing model weights are kept intact, and hence preserves existing capabilities, and (iii) Applies to diverse domains and settings.
We illustrate that augmenting \palms{} with a  smaller model trained on low-resource languages results in an absolute improvement of up to $13$\% on tasks like
translation into English and arithmetic reasoning for low-resource languages. 
Similarly, when  
\palms{} is augmented with a code-specific model, we see a relative improvement of $40$\% over the base 
model
for code generation and explanation tasks---on-par with fully fine-tuned counterparts.

\end{abstract}

\section{Introduction}

Large Language Models (LLMs) have shown to encompass a range of foundational capabilities such as 
commonsense and factual reasoning, world knowledge, and coherent language generation \citep{bubeck_sparks_2023, palm2}.
Leveraging these foundational capabilities, 
a number of efforts in the community have fine-tuned  these models to enable domain-specific capabilities such as code generation, copy editing, and mathematical problem solving \citep{lewkowycz_quant_2022, singhal_medpalm_2023}.
This has resulted in the development of several specialized large models with domain-specific capabilities.
For example, there are models that do well on standard code generation 
but are not as proficient in general logical reasoning and vice-versa.
Presence of such a large number of domain-specific models leads to a natural question:
Can we compose an \emph{anchor} model with a domain-specific \emph{augmenting} model to enable new capabilities? For example, can we compose an augmenting model's code understanding capability with an anchor LLM's language generation capability to enable code-to-text generation capability?

The typical approach for this problem is to further pre-train or (efficiently) fine-tune the anchor model
on the data that was originally used to train the augmenting model \citep{hu_lora_2021, kessler_adapter_2022}.
However, many a times such solutions are not feasible since
training large models is computationally expensive,
especially since the augmenting model itself may be an LLM
trained on a massive corpora. Further,
processing data from multiple sources
might not be feasible due to privacy concerns and organizational boundaries.
Working with multiple distinct models is also desirable
since it allows the reuse of  existing models with established capabilities,
providing better   control and avoiding catastrophic forgetting
that is prevalent in conventional approaches.

\begin{figure}[t]
    \centering
    \includegraphics[width=\textwidth]{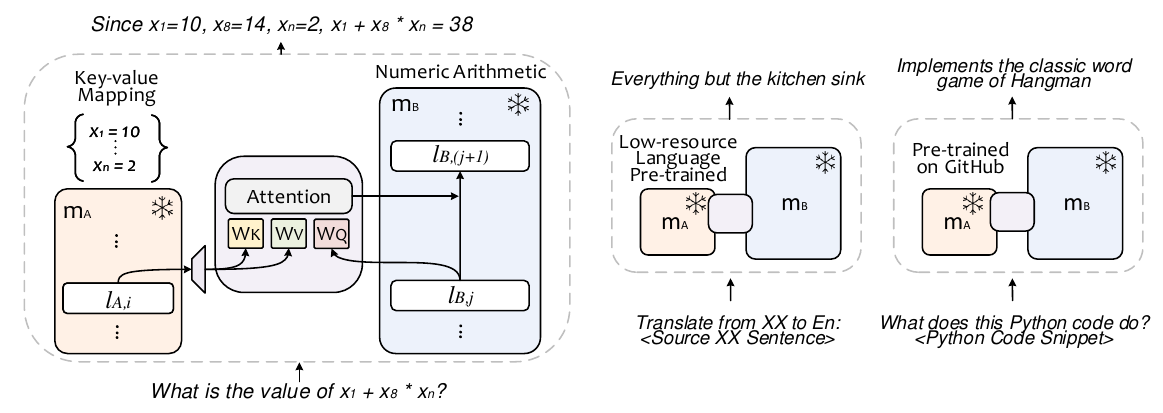}
    \emptysubfigure{sfig:kvpairs} %
    \emptysubfigure{sfig:approach} %
    \emptysubfigure{sfig:adapter} %
    \emptysubfigure{sfig:plots} %
    \caption{\label{fig:introduction} \textbf{Overview of \name{}.} To augment an \textit{\colorbox{largeColorT}{\textit{anchor}}}
    LLM (\anchorM) 
    with new capabilities 
    through \colorbox{composeColorT}{\textit{composition}} with a specialized \textit{\colorbox{smallColorT}{augmenting}} model  (\augM).
    Figure illustrates three \augM{}
    with different capabilities:
    key-value mapping (\textit{left}),
    low-resource languages (\textit{center}),
    and code (\textit{right}).
    Models \augM{} and \anchorM{}
    remain unchanged (\SnowflakeChevron) during composition. 
    A few additional parameters are learnt over models' layer representations. 
    Leftmost plot shows an \augM{} trained on a set of string-integer mappings, 
    e.g.,  \{$x_1: 10$, $\dots$, $x_n: 2$\}. \anchorM{} is a large LM with arithmetic capabilities. 
    \name{} composes these two frozen models to solve the task of arithmetic on keys which either models could not solve on their own  (\cref{sec:kv_experiments}). 
    Notably, \name{} generalizes to the entire key-value set despite training with arithmetic examples
    spanning only 20\% of the keys.
    }
\end{figure}

To address the training and the data challenges mentioned above,
we propose and study a practical setting for {\em model composition}:
(i) we are given access to one (or more) augmenting model(s)
and an anchor model, (ii) we are \emph{not allowed} to modify the weights of either models, and
(iii) we only have access to a small amount of data, representing the ``combined skills" of the given models, e.g., code generation with complex logical reasoning.

Prior work has largely approached the question 
of composition from either a routing or a merging standpoint, 
neither of which provide an effective solution to capture this setting.
Routing between the given models,
i.e., choosing an output of one model over the
other~\citep{ma2019snr}, or
performing a soft ensemble~\citep{softmerging} is not effective when neither of the models can demonstrate the desired capability.
Another body of work creates a combined model by  an  arithmetic
combination of base model parameters
~\citep{wortsman2022soups, taskarithmatic, fisheraverage}.
However, these settings are naturally restrictive
and their efficacy is unclear when
combining models with different sizes
and pre-training objectives~\citep{yadav2023resolving}.

In this work, we propose a novel {\textbf \fullName{}} ({\textbf \name{}}) framework
to address the general model composition setting mentioned above. 
Rather than a shallow combination of the augmenting and anchor LMs 
\citep{wortsman2022soups, taskarithmatic}, \name\
introduces a small number of  trainable parameters
over both augmenting and anchor models' intermediate layer representations.
 \name{} finds an effective
combination of the given models to perform new challenging tasks 
more accurately than either of the models alone,
while preserving the capabilities of individual models. 
Figure \ref{fig:introduction} highlights few motivating  scenarios for \name{}.

We study key practical applications of \name:
language inclusivity and code generation. 
For language inclusivity
(\cref{sec:language_experiments}),
we use a model that has been trained
on a set of low-resource languages.
We observe that composing this model
with the LLM allows us to borrow
its generation and reasoning capabilities
to achieve significantly better
performance on translation and
arithmetic reasoning tasks for 
low-resource languages
(Tables~\ref{tab:ntl-gatitos-results} and~\ref{tab:ntl-gsm-results}).
This composed model outperforms
not only the two base models but
also versions of the LLM that have been
further pre-trained or
LoRA \citep{hu_lora_2021} fine-tuned
for the set of low-resource languages.
For code generation (\cref{sec:code_experiments}),
we use a model that has been trained
on open-source code
across a variety of programming languages.
Composing this model with the
LLM---hence
borrowing its low-level logic
and generation capabilities---outperforms
the two base models
(Table~\ref{tab:code-results})
on code explanation and code completion tasks.

\section{Related Works}
\label{prior}

\paragraph{Parameter efficient fine-tuning:} A large body of work focuses on efficient ways of fine-tuning models for new domains by introducing a small number of trainable parameters, keeping the original model intact~\citep{houlsby2019adapter, wang_k-adapter_2020, pfeiffer_adapterfusion_2021, hu_lora_2021, kessler_adapter_2022}. Since this paradigm allows a small set of new parameters to be trained, it is challenging to use this approach to adapt a model to a new domain, which is absent from the original training corpus. 
In contrast, \name{} enables a model to be adapted to completely new domains using an augmenting model. In Section~\ref{sec:ablations}, we demonstrate that \name{} is significantly more effective than LoRA~\citep{hu_lora_2021}, a representative parameter efficient fine-tuning method.

\paragraph{Model Merging:} Merging different expert models with simple techniques like task vector averaging provides a way of recombining different capabilities of these models~\citep{taskarithmatic,fisheraverage}. 
However, these methods are only relevant when the original models are well aligned. Other related approaches are also applicable only when the models are derived from the same model~\citep{fisheraverage} or they are of same size ~\citep{softmerging}. In contrast, \name{} is more generic and is applicable to any set of models.

\paragraph{Model and Task Compositionality:} 
The modular encoder-decoder based method in  
\citep{dalmia_legonn_2022} adapts components of
encoder-decoder models to allow flexible
re-usability of different encoders, each
with their own capabilities.
Several past studies explore compositionality
from a multi-modal standpoint.
\cite{alayrac_flamingo_2022}
introduce cross-attention parameters across a language model
in order to attend to representations coming 
from an image encoder. They show very effective transfer
of capabilities between the two models.
In this work, we extend the ideology of model re-use and modularity to extend composition of capabilities in a large language model.

\paragraph{Models as Tools:}
Another interesting direction
for using multiple language models to solve a
downstream task has been to perform
composition in the models' input text space
\citep{zeng_socratic_2022, shen_hugginggpt_2023}.
\cite{schick2023toolformer} have demonstrated 
how a model can be taught to use external tools---there might be an opportunity to
investigate if other models can be
called as a part of the same framework.
Since these approaches require a large amount 
of prompt engineering,
in this work we focus on composition
through representations
that can be learnt automatically.

\section{\fullName{} (\name{})}
\label{sec:method}

Given an \textit{\textbf{anchor model}} \anchorM\ and an \textit{\textbf{augmenting model}} \augM, \name{} aims 
to compose the two models (\composed{\anchor}{\aug}) 
to enable new capabilities as a composition of capabilities of the two individual models.

As discussed in the introduction, we study this composition in a practical setting with the following assumptions: i) we can access weights, run forward and backward pass, and access intermediate representations of both \anchorM\ and \augM, ii) we are not allowed to change weights of both the models, iii) we do not have access to the training data, hyperparameters, training states of both the base models, iv) we are provided a few examples from the target composition domain.

The goal is to learn a composition
\composed{\anchor}{\aug} = $\mathnormal{f}$(\augM, \anchorM, $\Theta_{\mathbf{C}}$, \composeData{}) 
to achieve some joint task C. 
The weights of \augM\ and \anchorM\ are frozen. 
$\Theta_{\mathbf{C}}$ is the additional set
of trainable parameters
introduced to learn the composition
and
\composeData{} refers to the set of examples
that are used to learn this composition.

\subsection{\texorpdfstring{Learning to Compose ($\Theta_{\mathbf{C}}$)}{Learning to Compose}}
\label{sec:learning_compose}
As outlined in Figure~\ref{fig:introduction},
we operate over a selected set of layers
from \anchorM\ and \augM\ at all times.
We learn two sets of additional parameters
over these layers:
(i) A simple set of linear transformations,
$\projLayer$(.) that maps an
$i^{\text{th}}$ layer
representation from \augM\
to the dimensionality
of representations from \anchorM, and
(ii) A set of cross-attention layers,
$\crossAttn$(.,.)
that cross-attend between
this transformed layer representation
and a $j^{\text{th}}$
layer representation from \anchorM.

\textbf{Compositional Layers:}
Let the augmenting model \augM\
and the anchor model \anchorM\
have 
$N_{\aug}$ and $N_{\anchor}$
 layers, respectively. Also, let 
$D_{\aug}$ and $D_{\anchor}$
be the token 
dimensionality of the two models. 
We first choose a set of 
\textit{compositional} layers---\layerSet$_{\aug}$ and \layerSet$_{\anchor}$---for
both models, 
over which the set of new learnable
parameters are introduced during composition.
$n_{\aug} = |$\layerSet$_{\aug}|$ and 
$n_{\anchor} = |$\layerSet$_{\anchor}|$.
For simplicity, we set $n_{\aug}=n_{\anchor}=n$
and
the gap between two contiguous selected layers
is kept uniform based on the number
of selected layers---that is,
($l_{\layerIdxT{}{2}}-l_{\layerIdxT{}{1}}) =  \dots =\ (l_{\layerIdxT{}{n}} - l_{\layerIdxT{}{(n-1)}}) = N / n$.
Further, 
\repVectorSet$_{\aug}$ $\in \{\repVectorT_{\layerIdxT{\aug}{1}}, \repVectorT_{\layerIdxT{\aug}{2}}, \dots, \repVectorT_{\layerIdxT{\aug}{n_{\aug}}}\}$ denote the layer representation of a given input after each layer in \layerSet$_{\aug}$.

\textbf{Learned Projections:} Next we map representations from \augM\ to that of \anchorM\ via a projection layer. In particular, for each layer in  \layerSet$_{\aug}$, 
we learn a projection function 
$\projLayer: \mathbb{R}^{D_{\aug}}\rightarrow \mathbb{R}^{D_{\anchor}}$, 
that projects 
representations from these layers
to the desired representation size of \anchorM. Let, 
\begin{align*}
    \projLayer(\text{\repVectorSet}_{\aug}) &\xleftarrow{} \{\projLayerT{\aug}{1}, \projLayerT{\aug}{2}, \dots, \projLayerT{\aug}{n_{\aug}}\}
\end{align*}

This transformation enables cross-attention across models,
and also performs an alignment of representations from \augM\ and \anchorM\ despite frozen weights of the base models.

\textbf{Cross-attention Layers:}
Similar to the multi-headed cross-attention
in encoder-decoder models
(for example~\cite{vaswani2017attention} and~\cite{raffel2020t5})---we introduce cross-attention between representations of the anchor and the augmenting model. In particular, we use 
$\projLayerT{\aug}{i}$ from the
augmenting model
as the \textit{key} and \textit{value}
vectors for each head in cross-attention. We use the vector $\repVectorT_{\layerIdxT{\anchor}{j}}$
from the anchor model
as the \textit{query} vector, which leads to the following cross-attention setup:

\begin{align*}
    \crossAttnT{\aug}{i}{\anchor}{j}
    &= \text{Concat.}_k\left(\text{head}_{k}\right)\mathbf{W}^{O}\text{\quad}\forall k \in N_H \\
    \text{where,\ } \text{head}_{k}
    &= \text{Attn.}(\mathbf{Q}_{\anchor}, \mathbf{K}_{\aug}, \mathbf{V}_{\aug}),\\
    \text{and,\ } \mathbf{Q}_{\anchor} &= \repVectorT_{\layerIdxT{\anchor}{j}}\mathbf{W}_{k}^{Q}, \\
    \mathbf{K}_{\aug}, \mathbf{V}_{\aug} &= \projLayerT{\aug}{i}\mathbf{W}_{k}^{K},\ \projLayerT{\aug}{i}\mathbf{W}_{k}^{V}
\end{align*}

Here, $N_H$ represents the number of attention
heads used for cross-attention which, in our case,
is typically the same as the number of heads
used for self-attention in \anchorM.
Each of $\mathbf{W}^{O} \in \mathbb{R}^{D_{\anchor}\times D_{\anchor}}$, $\mathbf{W}_{k}^{Q}$, $\mathbf{W}_{k}^{K}$,
and $\mathbf{W}_{k}^{V}$ $\in$
$\mathbb{R}^{D_{\anchor}\times D_{\anchor}//N_H}$ 
are learnable weight matrices, where $k \in \{1..N_H\}$.

Finally,
the cross-attention output
is added as a residual connection
to the  layer representations of   \anchorM.
The resultant output vector, in-turn,
is the input to the succeeding layer in \anchorM:
\begin{align*}
    \repVectorT_{\layerIdxT{\aug}\oplus{\anchor}{j}} &= \repVectorT_{\layerIdxT{\anchor}{j}} + \crossAttnT{\aug}{i}{\anchor}{j}
\end{align*}
Here, $ \repVectorT_{\layerIdxT{\aug}\oplus{\anchor}{j}}$ denotes the input to the $(j+1)^{th}$ layer  of the composed model. 
All layers in \layerSet$_{\aug}$ and \layerSet$_{\anchor}$
are utilized in a similar manner.
Propagating over the remaining layers
in \anchorM\ gives us a final output token
$y_t$ decoded for the $t^{th}$ timestep.
Akin to usual auto-regressive decoding,
the output token for each time-step is appended
to the input: $x_{t+1} = x_t \oplus y_t$,
Since the updated input at each time step
is passed to both models,
all representations for the two models
are refreshed.

\subsection{\texorpdfstring{Composition Training Data (\composeData{})}{Composition Training Data}}
Since the target model \composed{\augM}{\anchorM}
involves a composition over the two models
\augM\ and \anchorM,
we construct 
the set of training examples \composeData{}
to depict a ``combined skill''
that enables $\Theta_{\mathbf{C}}$
to attend over the two models
appropriately for the target task.

Ideally, if the set of tasks
involved in composition task are
distinguished as $\mathbf{t}_{1}$
and $\mathbf{t}_{2}$
respectively,
then we design \composeData{}
to depict the a joint task $\mathbf{C}$.
For example, with respect to our
synthetic key-value setup:
our final task ($\mathbf{C}$) is to perform
arithmetic over a set of keys.
The augmenting model \augM\
is trained to learn the given key-value
pairs (notated as task, $\mathbf{t}_{1}$)
and the anchor model \anchorM\
is generic model that can perform numeric arithmetic well
(task $\mathbf{t}_{2}$).
For learning the set of parameters
$\Theta_{\mathbf{C}}$ for composition,
we consider \composeData{} to be
arithmetic over a held-in
set of keys (task  $\mathbf{C}$),
encompassing combined skills
from the two models.
In contrast to fine-tuning approaches
like LoRA \citep{hu_lora_2021}
that would require the entire
knowledge source (here, key-values)
during training time,
we find that training composition
on only a fraction
of the keys
can generalize
to the full set.

In other real world settings,
a clear distinction in specializing tasks for each model 
might be difficult to formulate and hence
defining a task that
captures the combined skills can be challenging.
We find that using a set of examples
that capture certain capabilities
of the two models suffices, i.e.,
some rough notion of
$\mathbf{t}_{\aug \cup \anchor}$.
For our language inclusivity task,
we use a mixture of examples containing
a 
small amount of 
low-resource language 
and
high-resource language data.

\paragraph{Composing multiple models:} Finally, we note that while the method has been presented for a setting with one anchor model and only one augmenting model, 
\name{} is applicable to multiple augmenting models as well. In particular, \name{} would require learning similar projection and cross-attention components between the anchor and each of the augmenting model. We leave a thorough investigation of this as a topic of future work.

\section{Experiments}
\label{sec:experiments}

We demonstrate the following in three domains:
\textbf{(a)} an anchor LLM (\anchorM{}) can be composed with an augmenting model (\augM{}) trained on mappings between string keys and number values to solve arithmetic expressions over those keys requiring both, knowledge of the KV mappings and arithmetic capabilities (\S\ref{sec:kv_experiments}); 
\textbf{(b)} how \name{} can be used to expand the language coverage of an anchor LLM (\anchorM) to low-resource languages it has not seen during pre-training. We show that an augmenting model (\augM) pre-trained on low-resource languages can be composed with such an anchor model to significantly improve translation and math-word problem solving capabilities in low-resource languages (\S\ref{sec:language_experiments}); 
\textbf{(c)} how code completion and explanation can be improved by composing an anchor LLM with an augmenting model (\augM{}) specializing in the code domain (\S\ref{sec:code_experiments}).

In all experiments, we start with a \palmxxs{} model and further train it on domain-specific data to arrive at an augmenting model (\augM{}) that is then kept frozen during composition.
Note that no task specific training data was used to train \name{}.
We use \palmxs{} or \palms{} models as the anchor LLM (\anchorM{}) that is also kept frozen during composition training.
For all our experiments, we set
$N_{\aug} / n = 4$,
i.e., we perform composition using every
$4$th layer output from \augM{}. 
Correspondingly, layers from \augM{} (\layerSet$_{\anchor}$)
are chosen such that $n_{\anchor} = n_{\aug} = n$,
hence
$n_{\anchor} = N_{\aug} / 4$.

\subsection{Key-value Arithmetic}
\label{sec:kv_experiments}
We first study the setting where we have
a small augmenting LM that has been trained to memorize string-to-integer key-value (KV) mappings, and
a large anchor LM that is capable of performing arithmetic over integers.
We wish to use \name{} to compose them and enable a new capability of solving arithmetic expressions containing those keys.

\paragraph{Key-Value Domain Knowledge} We first generate a repository of KV pairs containing $\text{N}_{\text{KV}} = 25$K pairs by sampling English strings of length $2-6$ characters from the vocabulary of the \palmxxs{} model and randomly assigning them unique integer values in the range $[1, \text{N}_{\text{KV}}]$. This constitutes the knowledge artifact, \knowledgeData{KV}. 
We further generate a collection of arithmetic expressions (\knowledgeData{KV-EXP}) containing addition ($+$), subtraction ($-$), and multiplication ($\times$) operations between $3-6$ keys by randomly sampling keys from \knowledgeData{KV} and operations to perform between them.

Using these arithmetic expressions, we generate three datasets:
\\[5pt]
\textbf{(i)} KV-Substitution (\knowledgeData{KV-SUBS}): This dataset maps each expression in \knowledgeData{KV-EXP}, to an expression where the keys are replaced by their corresponding values. For example, this dataset contains examples of the form ($\texttt{<K1>}+\texttt{<K2>}-\texttt{<K3>}$, $10 + 22 - 24$).
\\[5pt]
\textbf{(ii)} KV-Arithmetic (\knowledgeData{KV-MATH}): This dataset maps each expression in \knowledgeData{KV-EXP} to the numeric value arrived at by solving the arithmetic expression when the keys would be replaced by the corresponding values. For example, examples in this dataset look like ($\texttt{<K1>}+\texttt{<K2>}-\texttt{<K3>}$, $8$). 
\\[5pt]
\textbf{(iii)} Numeric-Arithmetic (\knowledgeData{NUM-MATH}): This dataset maps the value substituted version of each expression in \knowledgeData{KV-EXP} to the numeric value arrived at by solving the arithmetic expression. For example, examples in this dataset look like ($10 + 22 - 24$, $8$).
%
\paragraph{Models}
We obtain augmenting model \augM{} by further training a pre-trained \palmxxs{} model on \knowledgeData{KV-SUBS} to make it memorize the KV pairs in \knowledgeData{KV}. Note that, training on \knowledgeData{KV-SUBS} does not teach this augmenting model how to solve arithmetic expressions. 
Next, we use a pre-trained \palmxs{} model as the anchor model \anchorM. This model is capable of solving numeric expressions with decent performance (see Table~\ref{tab:kv-results}). Note that, this model has no knowledge of the KV pairs in \knowledgeData{KV}. 

We now take examples from the KV-Substitution dataset \knowledgeData{KV-SUBS} that only span $20\%$ of the keys in \knowledgeData{KV} to form the training data for composition (\composeData{}). We use \composeData{} to compose the augmenting model (\augM{}) having knowledge of \knowledgeData{KV} and the pre-trained anchor model \anchorM{} by training the composition parameters ($\Theta_{\mathbf{C}}$) using \name{} as explained in~\cref{sec:method}.
Both \augM{} and \anchorM{} are kept unchanged.

\paragraph{Evaluation Task} We evaluate the composed model \composed{}{} for its ability to solve arithmetic expressions containing keys from \knowledgeData{KV}. Specifically, we evaluate on the subset of \knowledgeData{KV-MATH} dataset that does not contain expressions used in \composeData{} during  training. This way, we are able to measure the composed model's ability to generalize to keys beyond what was observed during training.

\begin{wraptable}{r}{0.425\textwidth}
\begin{center}
\centering
\vspace{-0.70cm}
\resizebox{\linewidth}{!}{
\begin{tabular}{lrrr}
\toprule
                    & \multicolumn{1}{c}{\cellcolor{smallColorT}\begin{tabular}[c]{@{}c@{}}\augM \end{tabular}} &
                    \multicolumn{1}{c}{\cellcolor{largeColorT}\begin{tabular}[c]{@{}c@{}}\anchorM\end{tabular}} &
                    \multicolumn{1}{c}{\cellcolor{composeColorT}\begin{tabular}[c]{@{}c@{}}\name\\(\composed{KV}{LLM})\end{tabular}} \\ \midrule
\multicolumn{1}{l}{\knowledgeData{KV-SUBS}}    &   98.1    &       0.0       &        92.9       \\
\multicolumn{1}{l}{\knowledgeData{NUM-MATH}} &   4.2     &       73.7        &   72.0       \\ \midrule
\knowledgeData{KV-MATH}              & 0.7                              & 0.0                                             &     \textbf{84.3}        \\ 
                      \bottomrule
\end{tabular}}
\end{center}
\caption{
Evaluation (accuracy (\%)) for a synthetic key-value (KV) task.
\augM\ is trained to memorize the KV mappings 
while \anchorM\ excels at arithmetic 
We see that a composition \composed{KV}{LLM}
is able to perform arithmetic over held-out keys. 
\label{tab:kv-results}
}
\end{wraptable}

\paragraph{Results}
Table~\ref{tab:kv-results} shows the performance of the three models:
\augM, \anchorM, and \composed{\augM}{\anchorM} across the aforementioned datasets.
First, we observe that the augmenting model \augM{} achieves $98.1\%$ at the KV-Substitution task showing that memorizes \knowledgeData{KV} well. 
Next, we see that it performs poorly ($4.2\%$) at the Numeric-Arithmetic task showing that it does not have arithmetic capabilities.
As a result, this model is not able to solve arithmetic expressions containing keys from \knowledgeData{KV}.

As expected, the anchor model \anchorM{} gets $0\%$ accuracy on the KV-Substitution and KV-Arithmetic tasks as it has not seen any data from \knowledgeData{KV}.
However, it performs  well ($73.7\%$) on the Numeric-Arithmetic task demonstrating capability of arithmetic over numerals.

Lastly, we see that the composed model \composed{A}{B} 
is able to solve all tasks with high accuracy, especially the KV-Arithmetic task ($84.3\%$) which both the underlying models fail at.  This shows that the composed model is able to leverage the relevant capabilities from both the augmenting and anchor model to solve a complex task.

\subsection{Low-resource Language Inclusivity}
\label{sec:language_experiments}

\begin{table}[bh]
\centering
\begin{center}
\resizebox{\textwidth}{!}{\begin{tabular}{lrrrrrrrrrrr}
\toprule
                                                                                                                  & \multicolumn{11}{c}{FLORES-200 (XX to En; chrF1)}   \\ \cmidrule{2-12}
\multirow{-2}{*}{Model}                                                                                          & \multicolumn{1}{l}{lij} & \multicolumn{1}{l}{mr} & \multicolumn{1}{l}{taq} & \multicolumn{1}{l}{nn} & \multicolumn{1}{l}{su} & \multicolumn{1}{l}{ban} & \multicolumn{1}{l}{pl} & \multicolumn{1}{l}{th} & \multicolumn{1}{l}{min} & \multicolumn{1}{l}{acm} & \multicolumn{1}{l}{\textit{avg.}} \\ \midrule
\cellcolor{baseColorT}\palmxxs{}                                                                                 & 24.0                       & 16.5                       & 21.6                       & 33.3                      & 20.6                      & 2.1                       & 5.3                       & 63.2                      & 44.0                       & 59.8                        & 29.0                     \\
\multicolumn{1}{c}{\cellcolor{smallColorT}\begin{tabular}[c]{@{}r@{}}$+$ NTL (\augM)\end{tabular}}       & 32.0                       & 21.6                       & \underline{46.9}                       & 50.0                      & 40.6                      & 4.1                       & 4.0                       & 63.8                      & 47.8                       & 61.1                        & 37.2                     \\ \midrule
\cellcolor{largeColorT}\begin{tabular}[c]{@{}l@{}}\palms\ (\anchorM)\end{tabular} & \underline{32.6} & \underline{24.2} & 44.6 & \underline{50.8} & \underline{50.9} & \underline{5.4} & \underline{9.5} & \underline{69.0} & \underline{61.0} & \underline{68.6} & \underline{41.7} \\ \midrule
\cellcolor{composeColorT}\begin{tabular}[c]{@{}l@{}}\name\ (\composed{NTL}{LLM})\end{tabular} & \textbf{44.1} & \textbf{30.4} & \textbf{55.1} & \textbf{54.6} & \textbf{54.4} & \textbf{11.8} & \textbf{11.3} & \textbf{69.4} & \textbf{61.1} & \textbf{68.9} & \textbf{46.1} \\ \customdashline 
\rowcolor{dullGray}\multicolumn{1}{c}{\anchorM$+$NTL (\finetuned{\anchor}{NTL})}
& 48.1 & 39.1 & 59.2 & 57.5 & 57.3 & 11.4 & 9.9 & 69.4 & 61.4 & 69.0 & 48.2 \\ \bottomrule
\end{tabular}}
\end{center}
\caption{
Translation performance for XX to English direction on the FLORES-200 dataset~\citep{flores200}: 
We show results for a subset of 10 low-resource languages. Note that the composed model \composed{}{} significantly outperforms both \augM{} and \anchorM{}. On the complete language list, \composed{}{} outperforms both the underlying models for 175 of 192 languages (Appendix~\ref{sec:all_ntl_results}; Figure~~\ref{fig:flores_all_langs}).
\finetuned{\anchor}{NTL} represents a skyline where \anchorM{} has been further pre-trained on \knowledgeData{NTL}. The composed model achieves  similar performance for a tiny fraction of the training cost.
\label{tab:ntl-gatitos-results}
}
\end{table}

\setlength{\textfloatsep}{3pt}
In this section, we study if we can compose such a large anchor LM \anchorM{} with a smaller augmenting LM \augM{} that has been pre-trained on low-resource languages, to perform translation and math-word problem solving tasks presented in these low-resource languages.

\paragraph{Low-resource Language Corpora}
We use the long-tail language set and the associated corpora  from the Next Thousand Languages (NTL) effort~\citep{caswell2020language, bapna2022building} as the domain data \knowledgeData{NTL}.
This large-scale corpora contains web-crawled monolingual sentences and translation pairs for $\sim$1000 languages. The dataset has been used for language expansion in translation systems and language models~\citep{garcia2021harnessing, siddhant2201towards}.

\paragraph{Models}
Akin to~\cref{sec:kv_experiments}, we obtain augmenting model \augM{} by training the \palmxxs{} model on \knowledgeData{NTL} to impart knowledge about these low-resource languages to the model.
For \anchorM{}, we use the pre-trained \palms{} model. 
We use $\sim5\%$ of the same low-resource language corpora \knowledgeData{NTL} as the training data \composeData{} to compose \augM{} and \anchorM{} via \name{}. 
Since both models are untrained during composition, the anchor model \anchorM{} is \textit{not} trained on any of the low-resource language data.

\paragraph{Evaluation Tasks}
We evaluate the composed model \composed{}{} on two tasks:
\\[3pt]
\textbf{(i)} Translating text from a non-English language to English: We carry out these evaluations in a 5-shot in-context learning paradigm on the FLORES-200~\citep{flores200} dataset. This dataset contains examples for 200 high- and low-resource languages.
\\[3pt]
\textbf{(ii)} Performing grade school math word problems expressed in a non-English language: We evaluate on the multilingual version of the GSM-8K dataset~\citep{shi2023language} containing math word problems for English and 9 other high-resource languages. We further generated a silver-standard GSM-8K dataset for low-resource languages by automatically translating the English examples in GSM-8K to 25 low-resource languages supported by Google Translate.\footnote{
We perform quality evaluations in Appendix~\ref{tab:gsm8k_quality}.
}

\begin{table}[t]
\centering
\begin{subtable}{\textwidth}
\resizebox{\textwidth}{!}{
\begin{tabular}{lrrrrrrrrrrr}
\toprule
                                                                                                                 & \multicolumn{11}{c}{GSM8K (Low-resource Languages; Accuracy)} \\ \cmidrule{2-12} 
\multirow{-2}{*}{Model}                                                                                          & \multicolumn{1}{l}{meo} & \multicolumn{1}{l}{mfa} & \multicolumn{1}{l}{pcm} & \multicolumn{1}{l}{efi} & \multicolumn{1}{l}{min} & \multicolumn{1}{l}{ilo} & \multicolumn{1}{l}{ady} & \multicolumn{1}{l}{mai} & \multicolumn{1}{l}{nso} & \multicolumn{1}{l}{mzn}           & \multicolumn{1}{l}{\textit{avg.}} \\ \midrule
\cellcolor{baseColorT}\palmxxs{}                                                                                 & 5.2                     & 6.8                     & 6.8                     & 4.0                     & 5.6                     & 7.2                     & 6.0                     & 3.6                     & 7.2                     & \multicolumn{1}{r}{6.8}           & 5.9                      \\
\multicolumn{1}{c}{\cellcolor{smallColorT}\begin{tabular}[c]{@{}r@{}}$+$ NTL (\augM)\end{tabular}}       & 7.6                     & 4.0                     & 4.4                     & 3.2                     & 6.0                     & 4.8                     & 6.4                     & 3.2                     & 6.0                     & \multicolumn{1}{r}{4.8}           & 5.0                      \\ \midrule
\cellcolor{largeColorT}\begin{tabular}[c]{@{}l@{}}\palms{} (\anchorM)\end{tabular}          & 28.8                    & 14.0                    & \textbf{34.4}                    & \underline{14.8}                    & \textbf{25.2}           & \underline{14.8}                    & \underline{30.0}                    & \underline{22.8}                    & \underline{8.4}                     & \multicolumn{1}{r}{\underline{31.6}}          & 22.5                     \\ \midrule
\cellcolor{composeColorT}\begin{tabular}[c]{@{}l@{}}\name\ (\composed{NTL}{LLM})\end{tabular} & \textbf{34.0}           & \underline{17.6}                    & \underline{33.6}           & \textbf{18.0}           & 23.6                    & \textbf{16.8}           & \textbf{36.4}           & \textbf{24.8}           & \underline{8.4}                     & \multicolumn{1}{r}{\textbf{36.4}} & \textbf{25.0}            \\ \customdashline
\rowcolor{dullGray}\multicolumn{1}{l}{\finetuned{\anchor}{NTL}}                                                              & 33.2                    & 20.4           & 31.6                    & 14.0                    & 24.8                    & 14.0                    & 29.2                    & 21.2                    & 9.6            & \multicolumn{1}{r}{27.6}          & 22.6                     \\
\bottomrule
\end{tabular}}
\end{subtable}
\newline
\newline
\begin{subtable}{\textwidth}
\centering
\resizebox{\textwidth}{!}{
\begin{tabular}{lrrrrrrrrrrr}
\toprule
                                                                                                                  & \multicolumn{11}{c}{(High-resource Languages)}   \\ \cmidrule{2-12}
\multirow{-2}{*}{Model}                                                                                           & \multicolumn{1}{l}{\cellcolor{baseColorT}en} & \multicolumn{1}{l}{te} & \multicolumn{1}{l}{bn} & \multicolumn{1}{l}{sw} & \multicolumn{1}{l}{ja} & \multicolumn{1}{l}{zh} & \multicolumn{1}{l}{th} & \multicolumn{1}{l}{fr} & \multicolumn{1}{l}{es} & \multicolumn{1}{l}{de} & \multicolumn{1}{l}{\textit{avg.}} \\ \midrule
\cellcolor{baseColorT}\palmxxs{}                                                                                  & 5.6                                            & 4.0                    & 2.0                    & 7.6                    & 2.0                    & 4.4                    & 6.0                    & 6.8                    & 5.6                    & 9.2                    & 5.3                      \\
\multicolumn{1}{c}{\cellcolor{smallColorT}\begin{tabular}[c]{@{}r@{}}$+$ NTL (\augM)\end{tabular}}         & 4.8                                            & 3.6                    & 3.2                    & 4.8                    & \textbf{3.2}           & 7.6                    & 6.4                    & 9.2                    & 5.6                    & 7.2                    & 5.6                      \\ \midrule
\cellcolor{largeColorT}\begin{tabular}[c]{@{}l@{}}\palms{} (\anchorM)\end{tabular}           & \underline{36.8}                                           & \underline{19.2}                   & \underline{23.2}                   & \underline{16.0}                   & 2.0                    & \underline{39.2}                   & \underline{29.6}                   & \underline{38.0}                   & \underline{32.4}                   & \underline{43.2}                    & \underline{28.0}                     \\ \midrule
\cellcolor{composeColorT}\begin{tabular}[c]{@{}l@{}}\name\ (\composed{NTL}{LLM})\end{tabular} & \textbf{37.2}                                  & \textbf{28.0}          & \textbf{27.2}          & \textbf{18.0}          & \underline{2.4}                    & \textbf{43.6}          & \textbf{33.2}          & \textbf{42.8}          & \textbf{36.0}          & \textbf{49.2}                    & \textbf{31.8}            \\ \customdashline \rowcolor{dullGray}\multicolumn{1}{l}{\finetuned{\anchor}{NTL}}                                                                & 36.0                                           & 17.6                   & 18.4                   & 14.4                   & 0.8                    & 33.6                   & 27.2                   & 34.8                   & 31.2                   & 42.0                    & 25.6                     \\ \bottomrule
\end{tabular}
}
\end{subtable}
\caption{
Evaluations for grade-school mathematics (GSM)
problems on low-resource (LRL) and high-resource (HRL) languages.
We observe that \name\ yields significant gains for both evaluation sets.
Gains on the HRL set suggests that \name\ avoids catastrophic forgetting.
\label{tab:ntl-gsm-results}}
\vspace{0.5cm}
\end{table}

\paragraph{Results}
Table \ref{tab:ntl-gatitos-results} shows results on the
FLORES-200 dataset~\citep{flores200}, where the input
is a low-resource (XX) language sentence and the output
should be the corresponding English translation.
For 10 low-resource languages shown in the Table, we see that both the underlying models \augM{} and \anchorM{} are outperformed by our composed model \composed{}{}. 
We find that the composed model \composed{}{} outperforms \anchorM{} on 175 of the complete set of 192 languages (Appendix~\ref{sec:all_ntl_results}).

Table~\ref{tab:ntl-gsm-results} shows the performance of these models on the grade-school math word problems from the GSM8K task~\citep{cobbe2021gsm8k}
on low-resource languages (\textit{top}) and high-resource languages (\cite{shi2023language}; \textit{bottom}).
%
Firstly, we observe that the augmenting model \augM{} does not perform well on this task due to its limited mathematical reasoning capabilities. On the other hand, the anchor model \anchorM{} does much better given its mathematical reasoning capabilities and transfer-learning from high-resource languages. 
Finally, we observe that \composed{}{} outperforms both \augM{} and \anchorM{} on \textbf{18 of 25} low-resource and \textbf{9 of 10} high-resource languages, demonstrating effective composition of models. See Table~\ref{tab:gsm8k_complete} (Appendix~\ref{sec:gsm_8k_supp}) for a complete set of evaluations. Note that the last row in Table~\ref{tab:ntl-gsm-results} shows that  \anchorM{} when fine-tuned on  \knowledgeData{NTL} leads to worse performance than the pre-trained \anchorM{} indicating forgetting. Composing domain-specific model \augM{} with \anchorM{} using \name{} avoids this.

\subsection{Code Understanding and Generation}
\label{sec:code_experiments}

\begin{table}[t!]
\centering
\begin{center}
\resizebox{\textwidth}{!}{
\begin{tabular}{lr r rrrrrr}
\toprule
                                                                         Model &
                                            \multicolumn{1}{c}{CC (P@1)} &
                                              \multicolumn{1}{c}{T2C (P@1)}&     \multicolumn{6}{c}{C2T (chrF1)}   
                                              \\ 
                                              \cmidrule{2-9} 
                                           & \multicolumn{1}{c}{HumanEval} & \multicolumn{1}{c}{MBPP}                                           & \multicolumn{1}{l}{\cellcolor[HTML]{FFFFFF} Python} & \multicolumn{1}{l}{PHP} & \multicolumn{1}{l}{Go} & \multicolumn{1}{l}{Java} & \multicolumn{1}{c}{JS} & \multicolumn{1}{l}{Ruby}   
                                                                                      \\ \midrule
\multicolumn{1}{l}{\cellcolor{smallColorT}\begin{tabular}[c]{@{}l@{}}\palmxxs{} \\+ Code (\augM)\end{tabular}} &	19.5		&	28.0 & 28.0	&34.7	&32.6&	29.6&	26.5 &	26.0  \\
\multicolumn{1}{l}{\cellcolor{largeColorT}\begin{tabular}[l]{@{}l@{}}\palms{} (\anchorM)\end{tabular}}  &	16.4&		28.6   &  30.4 &	35.5&	40.4 &	31.0 &	28.8 &	27.9   \\
 \midrule
\cellcolor{composeColorT}\begin{tabular}[c]{@{}l@{}}\name{} (\composed{Code}{LLM})\end{tabular} &\textbf{22.5}		&	\textbf{32.2} &\textbf{30.5}&	\textbf{35.8}&	\textbf{40.6}&	\textbf{31.4}	&\textbf{29.3}	&\textbf{29.0}	\\

\customdashline \rowcolor{dullGray}\multicolumn{1}{l}{\finetuned{\anchor}{Code}}                                                                & 24.3                                                  & 43.0                   & 18.9                   & 35.0                   & 41.1                  & 31.1                   & 20.2                   & 27.6                                    \\

           \bottomrule
\end{tabular}}
\end{center}
\caption{
Evaluations for code generation 
and understanding across three tasks:
Code Completion (CC), Text-to-Code (T2C), and Code-to-Text (C2T).
Augmenting code understanding to \anchorM\ using \augM\
significantly improves performances across all datasets. \finetuned{\anchor}{Code} represents a skyline where \anchorM\ further pretrained on the \knowledgeData{Code}, which shows catastrophic forgetting of text generation task.            
\label{tab:code-results}
}
\vspace{0.5cm}
\end{table}

Code understanding and generation require two distinct types of capabilities: 
(\textbf{a}) knowledge of the syntax and semantics of code, 
and (\textbf{b}) knowledge of the world that the code is manipulating. 
While LLMs have a wealth of world knowledge,
they could often lack the specific knowledge of code syntax due to a skewed representation of code data in their pretraining corpora.
Conversely, small models trained specifically on code data
could exhibit a good understanding of code syntax,
but they may lack broad world knowledge and reasoning.
\name{}
can enable best of both worlds.

\paragraph{Code Domain Data}
Here, we use the code-specific corpus, \knowledgeData{Code},
consisting of open-source code extracted from GitHub heads for a variety of programming languages to train \augM{}.

\paragraph{Models}
Similar to~\cref{sec:kv_experiments}, 
a version of the \palmxxs{}
model has been further pre-trained
on \knowledgeData{Code} is used as
\augM, while the base pre-trained \palms{}
model acts as \anchorM.
We build \composed{\augM}{\anchorM} by training \name\ with only ~7\% of the same code data (data used for \augM) to have a data parity.

\paragraph{Evaluation Tasks}
We evaluate the efficacy of \name\ on three different tasks:
\\[5pt]
\textbf{(i)} Code-Completion (CC):
Given an initial set of lines of a code, the model is prompted to complete the code snippet. Here the aim is to evaluate the model for code syntax. We perform zero-shot evaluations on HumanEval benchmark dataset~\citep{humaneval} and report the Pass@1 (P@1) metric.\\[5pt]
\textbf{(ii)} Text-to-Code (T2C): Given a textual context, the model is prompted to generate the corresponding code snippet. Here, the evaluation indicates language understanding and code generation capabilities. We perform 3-shot inference on the MBPP dataset~\citep{mbpp} and report P@1.\\[5pt]
\textbf{(iii)} Code-to-Text (C2T):
Given a code snippet, the goal is to generate a natural language explanation of the code. This task evaluates code understanding and text generation. We perform 3-shot evaluations on the
CodeXGlue benchmark~\citep{codexglue} and report chrF1 scores across languages.

\paragraph{Results} 
Table~\ref{tab:code-results} reports comparative performance
for the individual models \augM{} and \anchorM{}, the composed version \composed{}{}, and a fine-tuned anchor baseline \finetuned{\anchor}{Code}.
Firstly, evaluations on the HumanEval dataset suggest
that \augM{} has a superior
understanding of code syntax as a result of its
additional training on \knowledgeData{Code}.
While, due to the larger scale and general purpose pre-training
of \anchorM{}, it excels at general language understanding and hence
performs better on the T2C and C2T tasks.
%

When employing \name{} to compose the two models,
we observe a clear transfer and composition of capabilities
through significant performance improvements:
$6.1\%$ and $3.6\%$ absolute gains over \anchorM{}
on the CC and T2C tasks, respectively.
%
We observe that fine-tuning \anchorM{} on \knowledgeData{Code} leads to a significant decline in the C2T performance due to catastrophic forgetting. \name{} retains the performance and is marginally better than \anchorM{} across all languages.
We also study qualitative examples on the C2T task
and observe interesting common patterns 
that are discussed in Appendix~\ref{sec:qualitative}.
%

\subsection{Ablations}
\label{sec:ablations}

\begin{table}[t]
\begin{center}
\begin{tabular}{llrrrrrr}
\toprule
                                                                              &                                                   & \cellcolor{baselineColorT}\finetuned{\anchor}{NTL/Code} & \begin{tabular}[c]{@{}l@{}}\cellcolor{composeColorT}\name{} \\ \cellcolor{composeColorT}\composed{NT}{LLM}\end{tabular} &
                                                                               \cellcolor{baselineColorT}\begin{tabular}[c]{@{}l@{}}Vanilla\\ \augM\end{tabular} & \cellcolor{baselineColorT}\begin{tabular}[c]{@{}l@{}}Random\\ \augM\end{tabular} & \cellcolor{baselineColorT}\begin{tabular}[c]{@{}l@{}}\augM as an\\ encoder\end{tabular} & \cellcolor{baselineColorT}\begin{tabular}[c]{@{}l@{}}LoRA\end{tabular} \\ \midrule
                                                                               & chrF1                                              & \textbf{62.1}                                                            & \underline{60.5}                                                               & 59.2                                                              & 58.8                                                                        & 59.3                                                                        & 59.2                                                                        \\
\multirow{-2}{*}{\begin{tabular}[c]{@{}l@{}}FLORES-200\\ (XX-En)\end{tabular}} & \#($>$\anchorM) & \underline{171}                                                           & \textbf{175}                                                   & 115                                                              & 43                                                                       & 102                                                                       & 82 \\ \midrule
                                                                               & Accuracy                                              & 19.8                                                       & \textbf{21.4}                                                          & 19.0                                                          & 17.8                                                                   & 19.1                                                                        & \underline{20.9} \\
\multirow{-2}{*}{\begin{tabular}[c]{@{}l@{}}GSM-8K\\ (LRL)\end{tabular}}       & \#($>$\anchorM) & \underline{15}                                                          & \textbf{20}                                                  & \underline{15}                                                            & 9                                                                      & 12                                                                      & \underline{15}   \\ \midrule
                                                                               & Accuracy                                             & 27.1                                                       & \textbf{33.1}                                                           & 29.7                                                          & 28.5                                                                   & 29.1                                                                        & \underline{31.2} \\
\multirow{-2}{*}{\begin{tabular}[c]{@{}l@{}}GSM-8K\\ (HRL)\end{tabular}}       & \#($>$\anchorM) & 1                                                          & \textbf{11}                                                  & 8                                                             & 4                                                                      & 6                                                                       & 
\underline{9} \\ 
\midrule
\midrule
HumanEval & Pass@1 &\textbf{24.3} & \underline{22.5} & 20.0 &20.1 &16.0 & 18.3 \\
\midrule
MBPP & Pass@1 &\textbf{43.0} & \underline{32.2}& 28.0& 27.0&27.0 & 28.7 \\
\midrule
CodeXGLUE & chrF1  & 29.0& \textbf{32.6}& 32.2& 32.1&32.0 & \textbf{32.6} \\
\bottomrule  
\end{tabular}
\end{center}
\caption{
Comparative performance of \name\
(\composed{NTL}{LLM})
across various possible ablations.
The metric ``\#($>$\anchorM)"
depicts the number of languages
for which the corresponding model
is better than the base for NTL, 
\anchorM---out of 192, 25, and 11
languages for the three tasks respectively.
For all compared settings,
the number of added parameters 
are kept the same.
\label{tab:ntl_ablations}
}
\end{table}

\paragraph{Influence of \augM}
We first study the influence
of 
\augM\ by replacing it
with vanilla and random variants
during composition.
Table~\ref{tab:ntl_ablations}
shows the variation of performance
across NTL and Code tasks
when the specialized \augM{} 
is replaced with a \textit{vanilla}
\palmxxs{} checkpoint
or an untrained version of the model,
i.e., a \textit{random} model.
We see that there is a considerable
drop of performance with these
variants across all tasks.
On FLORES-200 XX-En task,
languages improved with composition
drop to 115 and 43 with vanilla and random,
respectively.
A slight improvement of the vanilla
model over \anchorM{} indicates that
an un-specialized model
(with a different training regime than \anchorM{})
might have orthogonal capabilities
leading to an enhanced model.
This finding validates that performance gains
seen with \name{} is a result of utilizing \augM{}
and not the added $\Theta_{\mathbf{C}}$ parameters.

\paragraph{Influence of iterative decoding}
We also investigate a variation 
where we use \augM{} as an encoder,
i.e., an output token decoded at a given timestep
is not amended to \augM's
input. In this case, only the
prefix representations of \augM{} are used.
This setting eludes to past work
for image and text models \citep{alayrac_flamingo_2022}
where encoder and decoder models are composed.
We observe a significant decline in performance
across our various tasks when employing this setting.
%

\paragraph{Comparision with LoRA}
Finally, we evaluate a parameter efficient fine-tuning
approach by training LoRA~\citep{hu_lora_2021}
layers to adapt \anchorM.
For all experiments,
we set the LoRA rank such that
the number of added parameters is
equal to the number of parameters introduced with \name{}.
We also train LoRA on the same data as \name{}, i.e., \composeData{}.
We see a considerable difference in performance between
the two approaches across all tasks and metrics.


\section{Conclusion}
\vspace*{-5pt}

The proposed \name\ framework
composes an \textit{anchor}
LLM with specialized
\textit{augmenting} models to enable new tasks
not achievable by either models individually.
\name\ does not require updating the individual models
and learns a dense interaction between the models
through a few trainable cross-attention parameters.
Our experiments present consistent evidence that
\name\ learns to utilize the expertise
from the two models. That is, when composed with
relevant augmenting models, we observe a significant uptick 
in the anchor model's performance
across multiple challenging tasks, such as 
low-resource translation, reasoning,
and code explanation/generation.

That is, \name{}  is especially useful in scenarios where proprietary
data and knowledge is stored in parametric models.
With \name{}, a foundational LLM could be augmented with such proprietary models to extend a variety of foundational
capabilities such as reasoning, world knowledge, and coherent generation over the target proprietary domains.
Finally, extensions of \name{} could be used to acquire distinct knowledge from multiple augmenting models.  

\newpage
\section*{Acknowledgments}

This work was done during RB's pre-doctoral
tenure at Google Research, India (GRI)
with PT and PJ.
RB is indebted to Manish Gupta,
Divvy Thakkar, and all others
who enabled this oppurtunity.
RB would also like to thank the members
of the Languages team and other researchers at GRI
(and beyond),
including the incredible pre-doctoral cohort.
This work wouldn't have been possible without
their constant support.
Namely:
Aishwarya P.S., Laurent El Shafey, and Qiao Zhang
for their massive help in coding
and debugging;
Palak Jain and Sagar Gubbi for their feedback
and support throughout the project;
Kartikeya Badola, Shreyas Havaldar, Amandeep Kaur,
and Rishabh Tiwari for being the first ears
to all ideas;
Cyrus Rashtchian and Richa Dixit for their mentorship.

\bibliography{iclr2024_conference}
\bibliographystyle{iclr2024_conference}

\newpage
\appendix
\section{Supplementary Material for NTL}
\label{sec:all_ntl_results}
\subsection{FLORES-200}
\begin{figure}[h]
    \centering
    \includegraphics[width=\textwidth]{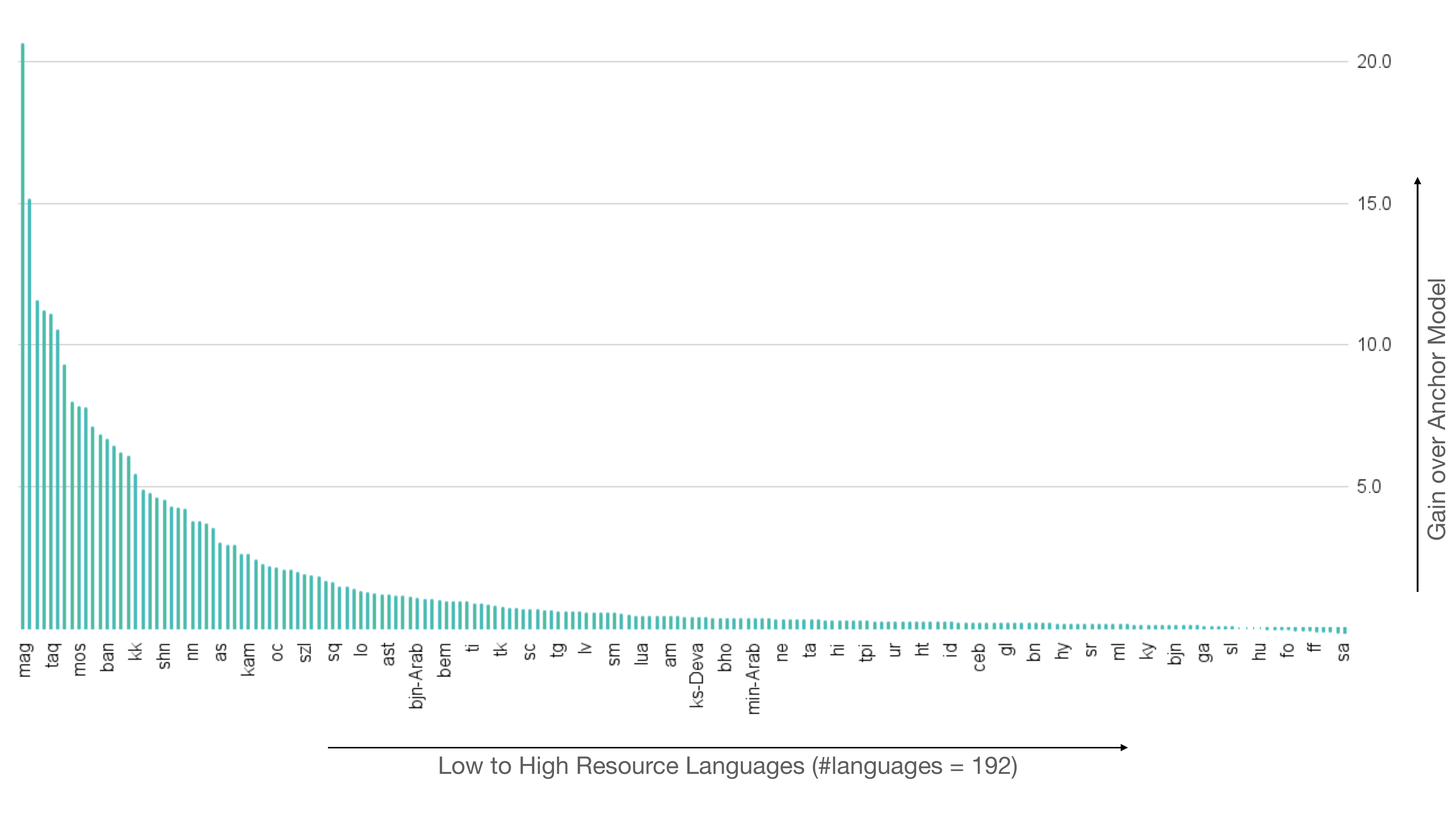}
    \caption{
    Gains seen by the composed model
    \composed{\anchor}{\aug} over the anchor
    model, \anchorM, for the complete set of FLORES-200 languages.
    The languages are sorted from
    low to high-resource.
    }
    \label{fig:flores_all_langs}
\end{figure}

Figure \ref{fig:flores_all_langs}
depicts the gains over the anchor
\palms{} model when augmented with a 
model that has been trained on \knowledgeData{NTL}.
We see a positive gain through \name\
for \textbf{175 of 192} languages.
The highest gains are seen for low-resource
languages since they are the most
underrepresented in the original model.
Diminishing returns with higher
resource languages is seen
and this trend is similar to the
trend seen for \finetuned{\anchor}{NTL}.

\begin{table}[ht]
\centering
\begin{subtable}{0.45\textwidth}
\centering
\resizebox{\textwidth}{!}{
\begin{tabular}{crrrr}
\toprule
 & \cellcolor{smallColorT}\augM & \cellcolor{largeColorT}\anchorM & \cellcolor{composeColorT}\begin{tabular}[c]{@{}l@{}}\composed{NTL}{LLM}\\ (\name)\end{tabular} & \cellcolor{gray!10}\finetuned{\anchor}{NTL} \\
\midrule
meo & 7.6 & \underline{28.8} & \textbf{34.0} & \cellcolor{gray!10}33.2 \\
mfa & 4.0 & \underline{14.0} & \textbf{17.6} & \cellcolor{gray!10}20.4 \\
pcm & 4.4 & \textbf{34.4} & \underline{33.6} & \cellcolor{gray!10}31.6 \\
efi & 3.2 & \underline{14.8} & \textbf{18.0} & \cellcolor{gray!10}14.0 \\
min & 6.0 & \textbf{25.2} & \underline{23.6} & \cellcolor{gray!10}24.8 \\
ilo & 4.8 & \underline{14.8} & \textbf{16.8} & \cellcolor{gray!10}14.0 \\
ady & 6.4 & \underline{30.0} & \textbf{36.4} & \cellcolor{gray!10}29.2 \\
mai & 3.2 & \underline{22.8} & \textbf{24.8} & \cellcolor{gray!10}21.2 \\
nso & 6.0 & \textbf{8.4}  & \textbf{8.4}  & \cellcolor{gray!10}9.6  \\
mzn & 4.8 & \underline{31.6} & \textbf{36.4} & \cellcolor{gray!10}27.6 \\
bew & 4.4 & \underline{33.6} & \textbf{34.8} & \cellcolor{gray!10}33.6 \\
ts  & 4.8 & \underline{7.2}  & \textbf{10.0} & \cellcolor{gray!10}11.6 \\
dv  & 2.8 & \underline{11.2} & \textbf{14.8} & \cellcolor{gray!10}13.2 \\
\bottomrule
\end{tabular}}
\end{subtable}
\hspace{0.05\textwidth}
\begin{subtable}{0.45\textwidth}
\centering
\resizebox{\textwidth}{!}{
\begin{tabular}{crrrr}
\toprule
 & \cellcolor{smallColorT}\augM & \cellcolor{largeColorT}\anchorM & \cellcolor{composeColorT}\begin{tabular}[c]{@{}l@{}}\composed{NTL}{LLM}\\ (\name)\end{tabular} & \cellcolor{gray!10}\finetuned{\anchor}{NTL} \\
\midrule
bho & 4.0 & \underline{23.6} & \textbf{29.2} & \cellcolor{gray!10}22.8 \\
cv  & 6.0 & \textbf{17.6} & \underline{16.4} & \cellcolor{gray!10}20.4 \\
mni & \underline{3.6} & 2.8  & \textbf{4.4}  & \cellcolor{gray!10}6.0  \\
or  & 2.4 & \underline{9.6}  & \textbf{12.4} & \cellcolor{gray!10}12.0 \\
kri & 5.6 & \underline{12.4} & \textbf{18.8} & \cellcolor{gray!10}20.0 \\
tk  & 5.2 & \underline{27.2} & \textbf{29.2} & \cellcolor{gray!10}28.8 \\
gom & 4.8 & \underline{22.4} & \textbf{25.2} & \cellcolor{gray!10}22.8 \\
ug  & 6.0 & \underline{23.2} & \textbf{29.2} & \cellcolor{gray!10}26.4 \\
ckb & 3.2 & \underline{25.6} & \textbf{28.0} & \cellcolor{gray!10}27.2 \\
as  & 1.2 & \underline{5.2}  & \textbf{9.2}  & \cellcolor{gray!10}4.0  \\
doi & 3.6 & \underline{17.2} & \textbf{22.4} & \cellcolor{gray!10}21.6 \\
dz  & \textbf{4.4} & \underline{0.8}  & 0.4  & \cellcolor{gray!10}0.0  \\ \midrule
avg.& 4.5 & \underline{18.6} & \textbf{21.4} & \cellcolor{gray!10}19.8 \\
\bottomrule
\end{tabular}}
\end{subtable}
\caption{
Performance evaluations on the complete set of
low-resource languages for GSM-8K.
Augmenting \augM\ with \anchorM\ as \composed{NTL}{LLM}
improves performance over \anchorM\ across a majority
of languages. On average, we see an improvement of 2.8\%.
\label{tab:gsm8k_complete}
}
\end{table}

\begin{table}[ht]
\centering
\begin{subtable}{\textwidth}
\centering
\begin{tabular}{crrrrrrrrrrrrrr}
\hline
\rowcolor{gray!30}
 & meo & mfa & pcm & efi & min & ilo & ady \\
\hline
Overlap & 83.17 & 75.54 & 81.28 & 78.35 & 77.90 & 77.80 & 76.21 \\
\hline
Delta & 1.15 & 1.25 & 1.18 & 1.22 & 1.23 & 1.24 & 1.28 \\
\hline
\end{tabular}
\end{subtable}
\hspace{0.05\textwidth}
\begin{subtable}{\textwidth}
\centering
\begin{tabular}{crrrrrrrrrrrrrr}
\hline
\rowcolor{gray!30}
 & mai & nso & mzn & bew & ts  & dv  & bho \\
\hline
Overlap & 76.63 & 69.58 & 71.32 & 71.37 & 61.62 & 55.18 & 73.67 \\
\hline
Delta & 1.26 & 1.40 & 1.38 & 1.37 & 1.55 & 1.70 & 1.30 \\
\hline
\end{tabular}
\end{subtable}
\hspace{0.05\textwidth}
\begin{subtable}{\textwidth}
\centering
\begin{tabular}{crrrrrrrrrrrrrr}
\hline
\rowcolor{gray!30}
 & cv  & mni & or  & kri & tk  & gom & ug  \\
\hline
Overlap & 58.52 & 58.94 & 68.03 & 77.18 & 66.06 & 71.21 & 57.66 \\
\hline
Delta & 1.62 & 1.60 & 1.45 & 1.27 & 1.48 & 1.36 & 1.65 \\
\hline
\end{tabular}
\end{subtable}
\caption{
Quality evaluation for the
LRL GSM-8K dataset across languages.
We created the dataset by translating the original
English sentences of GSM-8K to the target
language using the Google Translate API.
We measure quality by back-translating
the obtained examples back to English and measuring:
(i) The \textit{\textbf{overlap}} between the back-translated and the
original English sentence, and
(ii) The \textit{\textbf{delta}} change in performance when \palms{}
is evaluated on this back-translated version
of GSM-8K as compared to the original version.
\label{tab:gsm8k_quality}
}
\vspace*{10pt}
\end{table}

\subsection{GSM-8K}
\label{sec:gsm_8k_supp}
\paragraph{Quality evaluation for LRL GSM-8K}
As described in Section~\ref{sec:language_experiments},
we created the GSM-8K dataset~\citep{cobbe2021gsm8k}
for low-resource languages by using the Google Translate API
to obtain silver translations in the target language
from the source English sentence in the original dataset.
We perform a quality evaluation of these examples
by back-translating them back to English using the same
translation API and defining two metrics over it:\\
(i) Overlap:
The BLUE score measure between the actual example
and the back-translated example, \\
(ii) Delta:
The change in performance of the \palms{} model
when evaluated on the original GSM-8K set
as compared to the back-translated version.

Table~\ref{tab:gsm8k_quality} shows the values
for these metrics across the various languages.
We see that a decently high overlap value is seen
across all languages. At the same time,
the delta in performance is also minimal
indicating that key attributes in the
GSM-8K examples are not affected by translation.

\paragraph{Results on the complete language set}
Table~\ref{tab:gsm8k_complete} shows the comparative
evaluations on the complete set of 25 low-resource languages
for which GSM evaluations are performed.
We see an improvement over the anchor model \anchorM\
for \textbf{20 of 25} languages. We also compare against the fully
continued pre-trained version \finetuned{\anchor}{NTL}
and observe that \composed{\aug}{\anchor}
outperform it for \textbf{18 of 25} languages.

\section{Qualitative Analysis}
\label{sec:qualitative}
\begin{table}[thp]
\begin{adjustbox}{max width=\textwidth}
\begin{tabular}{cc}
\toprule
\begin{minipage}{0.5\textwidth}
\begin{tabular}{l}
\cellcolor{dullGray}
\begin{tabular}[l]{@{}l@{}}
\begin{python}
def ConsumeBool(self):
  try : 
    result = ParseBool(self.token) 
  except ValueError as e :
    raise self._ParseError(str(e)) 
  self.NextToken() 
  return result
\end{python}
\\ \\
\end{tabular} \\
    \begin{tabular}[c]{@{}p{0.9\textwidth}@{}}$\Rightarrow$ Consumes a boolean \end{tabular}
    \\
    \cellcolor{dullGreen}
    \textcolor{smallColor}{\textbf{\augM}}:\quad
    \begin{tabular}[c]{@{}l@{}}Consumes a boolean
    \end{tabular}
    \\
    \cellcolor{dullRed}
    \textcolor{largeColor}{\textbf{\anchorM}}:\quad
    \begin{tabular}[c]{@{}l@{}}The object is not a member
    \end{tabular}
    \\
    \cellcolor{dullGreen}
    \textcolor{composeColor}{\name}:\quad
    \begin{tabular}[c]{@{}l@{}}Consumes a boolean
    \end{tabular}
    \\
\end{tabular} \\
\end{minipage} &
\begin{minipage}{0.5\textwidth}
\begin{tabular}{l}
\cellcolor{dullGray}
\begin{tabular}[l]{@{}l@{}}
\begin{python}
def value(self): 
  if self.has_value:
    return self._impl[OBJ].get_val(K)
  else:
    raise ValueError("Not found")
  return
\end{python}
\\ \\ \\
\end{tabular} \\
\begin{tabular}[c]{@{}l@{}}$\Rightarrow$ Print an error message and exit. \end{tabular}
\\
\cellcolor{dullRed} 
\begin{tabular}[c]{@{}l@{}}
[a part of the given model prefix] \end{tabular}
\\
\cellcolor{dullGreen} 
\begin{tabular}[c]{@{}l@{}}
Exit with error message \end{tabular}
\\
\cellcolor{dullGreen} 
\begin{tabular}[c]{@{}l@{}}
Print an error message and exit \end{tabular}
\\
\end{tabular} \\
\end{minipage} \\ \midrule
\begin{minipage}{0.5\textwidth}
\begin{tabular}{l}
\cellcolor{dullGray}
\begin{tabular}[l]{@{}l@{}}

\begin{python}
def get_positions(url):
  data = _get_resource(url)
  positions = [x for x in data['p']]
  return positions
\end{python}
\\ \\
\end{tabular} \\
\begin{tabular}[c]{@{}l@{}}$\Rightarrow$ Returns a list of positions.\end{tabular} \\
\cellcolor{dullRed}
\begin{tabular}[c]{@{}l@{}}
Positions of specified instruments. \end{tabular}
\\
\cellcolor{dullRed}
\begin{tabular}[c]{@{}l@{}}
Get all positions.
\end{tabular}
\\
\cellcolor{dullGreen} 
\begin{tabular}[c]{@{}l@{}}Returns a list of positions . \end{tabular}
\\
\end{tabular}
\end{minipage} &
\begin{minipage}{0.5\textwidth}
\begin{tabular}{l}
\cellcolor{dullGray}
\begin{tabular}[l]{@{}l@{}}
\begin{python}
def distance(x0, y0, x1, y1):
  return (
    sqrt(pow(x1-x0,2) + pow(y1-y0,2)
  )
\end{python}
\\ \\
\end{tabular} \\
\begin{tabular}[c]{@{}p{0.925\textwidth}@{}}$\Rightarrow$ Returns the distance between two points \end{tabular}
\\
\cellcolor{dullGreen} 
\begin{tabular}[c]{@{}l@{}} Calculates the distance between two points \end{tabular}
\\
\cellcolor{dullGreen} 
\begin{tabular}[c]{@{}l@{}} Return the distance between two points \end{tabular}
\\
\cellcolor{dullGreen} 
\begin{tabular}[c]{@{}l@{}}Calculates the distance between two points\end{tabular}
\\
\end{tabular} \\
\end{minipage} \\
\bottomrule
\end{tabular}
\end{adjustbox}
\caption{
Cherry-picked qualitative examples for the code-to-text task on Python
that depict examples that fall
into a set of larger bucket of patterns
that we observe across examples.
\name\ does well in various
settings:
(i) when \augM
produces the correct output
but not \anchorM,
(ii) vice-versa---when
\anchorM\ does well,
and (iii) when neither of the two
base models do well but a combination
of intermediate representations allow
the composed model to give the correct
output.
This shows that composition
implicitly learns to do both:
routing across models and
a combination, based on a given input.
\label{tab:code_qual}
}
\vspace{5pt}
\end{table}

Table~\ref{tab:code_qual} depicts 
a few qualitative examples
for the code-to-text, or the code explanation
task, for Python.
These examples depict examples for the 
three broader bucket of examples
that we observe in cases when \name\
yields the correct responses:
\begin{enumerate}
    \item When neither of \augM\ or \anchorM\
generates the correct response but \composed{\aug}{\anchor} correctly
attends over their latent representations
to yield the correct output,
    \item When either of \augM\ or \anchorM\
is seen to give the correct response
while the other one is incorrect
and \composed{\aug}{\anchor} generates
the correct response that matches the 
generation from the correct model of
\augM\ and \anchorM, and
    \item When both \augM\ and \anchorM\
generate the correct response and 
\composed{\aug}{\anchor} 
reproduces those generations.
\end{enumerate}

We also observed similar qualitative patterns with
other tasks for language inclusivity.

\section{Overhead with \name{}}
In this section, we include a detailed computation of the expected parametric and training overhead while composing given models using our proposed \name{} framework.

\subsection{Parametric Overhead}
Building from the notations in \cref{sec:learning_compose}, let's say the two models \augM{} and \anchorM{} have $N_{\aug}$ and $N_{\anchor}$ number of standard transformer layers, respectively, with each layer of output dimensionality $D_{\aug}$ and $D_{\anchor}$. As mentioned, we choose $n = n_{\aug} = n_{\anchor}$ number of layers to perform the composition.
\begin{align*}
    \text{\# Parameters for each $\projLayer$ layer} &= (D_{\aug} * D_{\anchor}) \\
    \text{\# Parameters for each $\crossAttn$ layer} &= (3 * D_{\anchor}^2) \\
    \text{\# Parameters added during composition} &= n * (D_{\aug} * D_{\anchor} + 3 * D_{\anchor}^2) \\
    \text{\# Parameters in \anchorM} &= N_{\anchor} * (V_{\anchor}*D_{\anchor} + 3 * D_{\anchor}^2 + 2*D_{\anchor}*D_{\anchor}*K_{\anchor})
\end{align*}
\text{where, $V_{\anchor}$ and $K_{\anchor}$ depict the vocabulary size and hidden multiplication factor, respectively.}

Let's consider some standard transformer configurations to understand the parameter overhead.
As an example, consider the layer configurations of standard BERT models: BERT-small (\augM) and BERT-large (\anchorM). In this case: $N_{\aug}$ = 4, $D_{\aug}$ = 512, $N_{\anchor}$ = 24, $D_{\anchor}$ = 1024, $V_{\anchor}$ = 30K, $K_{\anchor}$ = 4. Assuming that we select all layers of \anchorM, the value of $n$ = 4.
Hence,
\begin{align*}
    \text{\# Parameters added during composition} &= 4  * (512 * 1024 + 3*1024^2) \approx  1.5\times10^7 \approx 15\text{M}\\
    \text{\# Parameters in \anchorM} &= 24 * (30\text{K}*1024 + 3 * 1024^2 + 2*1024^2*4) \approx 1\text{B}\\
    \text{\%age of new parameters added} &= 15\text{M} * 100 / 1\text{B} = 1.5\%
\end{align*}
\textbf{Hence, number of parameters added during composition $\approx$ 1.5\% of those in \anchorM.}

\subsection{Training Overhead}
While back propagation over \anchorM{} is indeed required while training CALM, the total training costs are still significantly lesser than training \anchorM{}, owing to the training examples/iterations required.

Firstly, as discussed above, the additional number of parameters introduced during composition is ~1.5\% of the number of parameters of \anchorM{}—hence, a negligible parametric addition. 

Further, since only ~5-7\% of the total \anchorM{} fine-tuning data is required to train CALM, the training cost of CALM is minimal with respect to training cost of training the entire anchor model.

Moreover, since our experiments consider an \augM{} that has 5-20\% of parameters as \anchorM{}, even the net cost of training \augM{} and CALM is significantly lesser than training \anchorM{}. 

Let’s assume that (i) the cost of fine-tuning \anchorM{} on the complete data is $X$, (ii) number of parameters in \augM{} is 10\% of those in \anchorM{}, and (iii) the amount of data required to train CALM is 2\% of \anchorM{} training. Assuming a linear scaling factor of training cost (FLOPS) with model parameters and data:

\textbf{Cost of training CALM $\approx$ $0.02 \times X$ = 2\% of \anchorM{} training. \\
Cost of training \augM{} + CALM $\approx$ $(0.10*X + 0.02*X) = 0.12\times X$ = 12\% of \anchorM{} training.}

\end{document}